# Robust deep labeling of radiological emphysema subtypes using squeeze and excitation convolutional neural networks: The MESA Lung and SPIROMICS Studies

*Artur Wysoczanski, *Nabil Ettehadi, Soroush Arabshahi, Yifei Sun, Karen Hinkley Stukovsky, Karol E. Watson, MeiLan K. Han, Erin D Michos, Alejandro P. Comellas, Eric A. Hoffman, Andrew F. Laine, R. Graham Barr, and Elsa D. Angelini

*Abstract*—Pulmonary emphysema, the progressive, irreversible loss of lung tissue, is conventionally categorized into three subtypes identifiable on pathology and on lung computed tomography (CT) images. Recent work has led to the unsupervised learning of ten spatially-informed lung texture patterns (sLTPs) on lung CT, representing distinct patterns of emphysematous lung parenchyma based on both textural appearance and spatial location within the lung, and which aggregate into 6 robust and reproducible CT Emphysema Subtypes (CTES). Existing methods for sLTP segmentation, however, are slow and highly sensitive to changes in CT acquisition protocol. In this work, we present a robust 3-D squeeze-and-excitation CNN for supervised classification of sLTPs and CTES on lung CT. Our results demonstrate that this model achieves accurate and reproducible sLTP segmentation on lung CTscans, across two independent cohorts and independently of scanner manufacturer and model.

*Index Terms*—Emphysema, squeeze and excitation, deep learning, MESA, SPIROMICS

## I. INTRODUCTION

PULMONARY emphysema is defined as an irreversible loss of lung tissue unrelated to fibrosis, and is characterized on pathology by enlargement of airspaces with annihilation of alveolar walls [1]; emphysema and chronic obstructive pulmonary disease (COPD), together, are the third leading cause of death globally [2]. Emphysema has traditionally been subdivided into three subtypes identified on pathology, namely centrilobular (CLE), panlobular (PLE), and paraseptal (PSE) [3, 4] which have shown associations with different risk factors and clinical manifestations [5, 6], and remain in widespread clinical use. However, these traditional subtypes were identified on autopsy series with quite limited sample size, and considerable overlap exists in radiographic appearance between the subtypes which limits radiologist inter-rater reliability [7]. Recent work has introduced quantitative descriptors of lung texture on CT which outperform radiologists on classification of centrilobular, panlobular and paraseptal emphysema [8, 9], and furthermore identify novel spatially-informed lung texture patterns (sLTPs) on full lung and cardiac CT scans [9-13]. Clinical visual inspection of these 10 learned sLTPs, along with utilization of statistical data reduction techniques, have further yielded a total of six CT emphysema subtypes (CTES), with distinct correlates to pulmonary functional measures, genetic variants and clinical outcomes that suggest they represent distinct disease subphenotypes [10].

A variety of artificial intelligence approaches have been developed for detection and classification of CLE, PLE and PSE [9, 14-18], in addition to CTES segmentation on both full-lung and cardiac CT [11, 13]; however, these approaches for segmentation of emphysema subtypes on CT are not necessarily robust to to changes in CT acquisition protocol or across different cohorts. In this paper, we introduce a deep learning (DL) framework consisting of a 3-D squeeze-and-excitation convolutional neural network (SE-CNN) model for classification of the 10 sLTPs and 6 CTES. Our model is trained, validated, and tested in inspiratory CT scans of the SubPopulations and Intermediate Outcome Measures in COPD Study (SPIROMICS) [19, 20], and we further confirm our model's cross-cohort generalizability by evaluating its performance on inspiratory CT scans acquired in Exam 5 of the Multi-Ethnic Study of Atherosclerosis (MESA) [21].We additionally test the scan-rescan reproducibility of deep-learned sLTP labels in the SPIROMICS Repeatability Study [22].

Our manuscript is organized as following: the network architecture and implementation are detailed in section II. Section III details the CT datasets used in the study, pre-processing, and sample generation. Our experimental results, including the performances of our model on both datasets, relative performance across scanner models, and scan-rescan reproducibility, are discussed in section IV. The discussion is presented in section V. Finally, section VI summarizes the conclusions.

## II. NETWORK ARCHITECTURE AND TRAINING

Our model takes as input a cubic region of interest (ROI) of size 36x36x36 voxels, with 1 channel, and returns a length-10 vector of class probabilities for each sLTP label. The model architecture is composed primarily of a series of 1) residual blocks, and 2) Squeeze and Excitation (SE) blocks [23]. Residual blocks (Figure 2a) consist of two parallel branches; the left branch consists two 3D convolutional layers with N channels. The first convolutional layer is followed by batch normalization and ReLU activation. The second convolutional layer involves a batch normalization layer only. The right branch involves a single 3D convolutional layer with N channels followed by a batch normalization layer. Finally, the outputs of the two branches are added together and followed by a ReLU activation. SE blocks (Figure 2b) are a form of self-attention units for CNNs introduced by [23], which selectively amplifies or suppresses the output channels of a convolutional layer, dependent on their relevance to the classification task. The SE block consists of two branches: a) a skip connection branch (left), and b) a branch that finds the optimal channel representation for the classification task. In the latter, a 3D global average pooling operation isolates the mean activation of each input channel, while two fully-connected layers (downsampling and then upsampling the number of channels by a tunable factor *r*) and a sigmoid activation learn the optimal channel weights. Channel weights are then multiplied with the input tensor at the end of the skip-connection. The residual and SE blocks are combined together to form a single block named *residual block with SE* (Figure 2c). The final deep learning model is consisted of 4 stacked *residual blocks with SE* (Figure 2d). The image dimension is reduced twofold after each block by 3-D max pooling. After the last 3D Maxpooling layer, the features are unrolled and fed through two fully connected layers with ReLU activation, with 512 and 128 nodes, respectively. Dropout with probability 0.5 was added to each fully-connected layer to prevent overfitting. Finally, a softmax layer with 10 classes is used to generate the output sLTP classification. For all 3D convolutional layers, kernels of size [1,1,1] with no paddings were used to capture the fine details of input ROI structure. The number of kernels for convolutional layers starts at 64 and doubled through each residual block with SE to reach 512 at the final block. The channel reduction factor *r* was set to 16 for all blocks. All 3D max-pooling layers used kernels of size [2,2,2] with no padding to halve the size of feature maps in each dimension between successive convolutional blocks. The model has a total of 2,914,890 parameters of which 5,760 are non-trainable.

The SE CNN sLTP classifier was implemented in Python using Keras with TensorFlow as backend on an NVidia 2080 ti RTX (Titan RTX) GPU with 24GB memory. Using a batch size of 32, the model was trained for 350 epochs with the categorical cross-entropy loss function. Model training was performed using Stochastic Gradient Descent (SGD) with Nesterov optimizer [24] (initial learning rate = 0.0001, Momentum = 0.6, and decay rate = $10^{-6}$). Although ADAM optimizer [25] has shown very promising results in training deep learning models faster, it does suffer from generalizability issues compared to SGD[26]. We initially trained our model using ADAM, but observed poor generalization to the test set. Hence in-line with claims made in [26] about the generalizability power of SGD, we chose SGD in this work and sped up the training process by using momentum. Figure 5 depicts the training history (model loss and overall accuracy as functions of epochs) of our model. The best performing model was chosen as the model providing the highest accuracy on the validation set.

### III. DATA AND PRE-PROCESSING

SPIROMICS is a prospective, longitudinal COPD case-

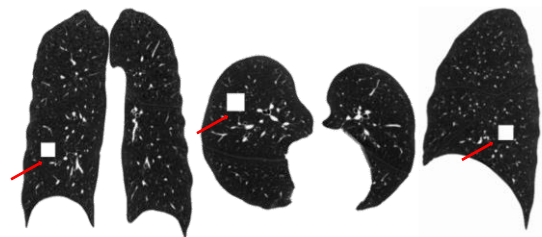

Figure 1. Example of an arbitrary 36x36x36 voxel ROI in the lungs of a representative SPIROMICS participant (left, coronal; center, axial; right, sagittal). The ROI is colored in white and is shown by the red arrows.

control study of participants ages 40-80 years recruited in 2010-2015, at 12 clinical centers in Winston-Salem, NC; Ann Arbor, MI; San Francisco, CA; Los Angeles, CA; New York City, NY; Salt Lake City, UT; Iowa City, IA; Baltimore, MD; Denver, CO; Philadelphia, PA; Birmingham, AL; and Chicago, IL. Participants included (1) smokers with spirometry-defined COPD and a history of over 20 pack-years, (N=2,100), (2) smokers with a history of over 20 pack-years and no COPD (N-

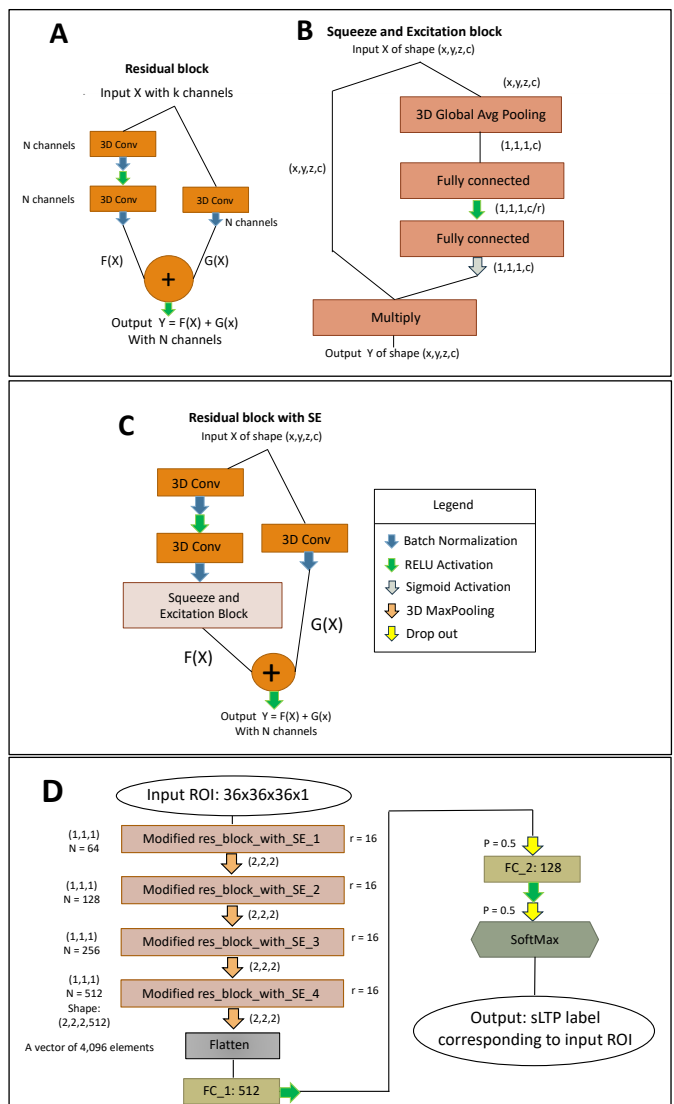

Figure 2. Schematic diagram of the SE-CNN model architecture and its constituent blocks: (A) residual block, (B) squeeze and excitation block, (C) residual block with SE, and (D) the SE CNN sLTP classifier

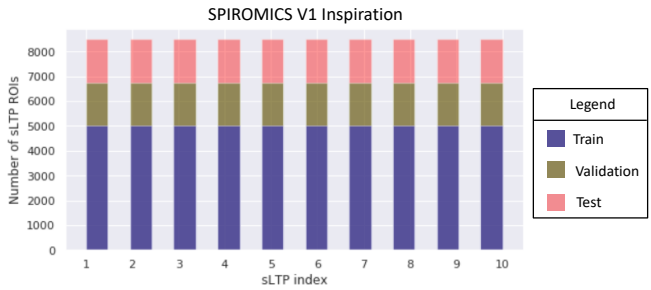

Figure 3. Balanced distribution of the sampled 3D ROIs from the SPIROMICS dataset

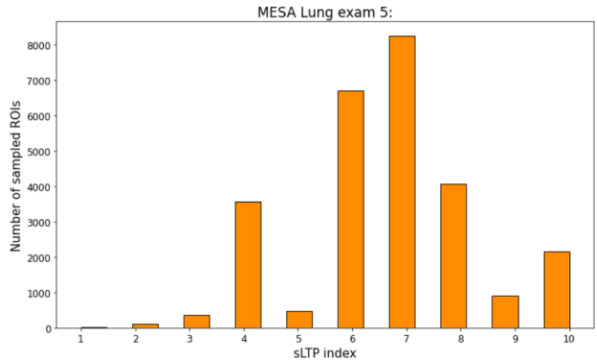

Figure 4. Distribution of the sampled 3D ROIs from the MESA Lung exam 5 dataset used to test model generalizability across cohorts.

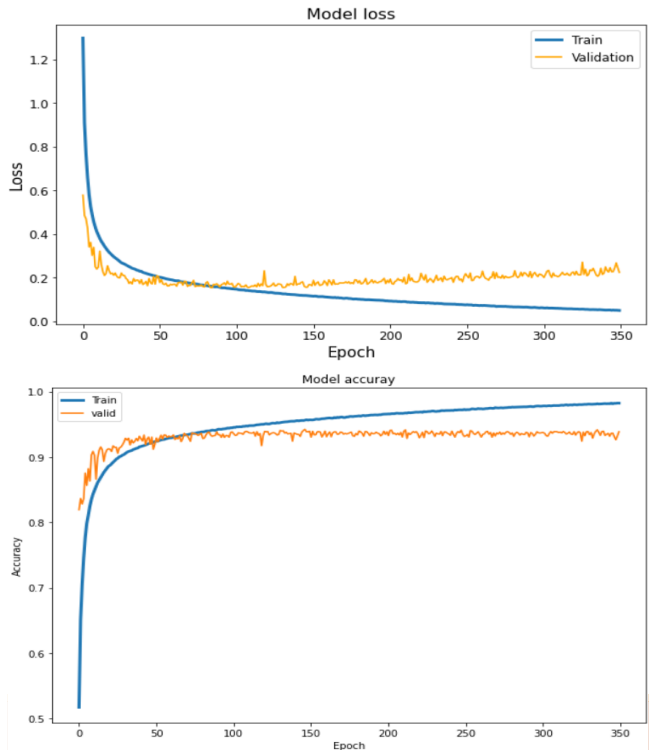

Figure 5. Training history of the SE CNN sLTP classifier. Loss (top) for train and validation sets (SPIROMICS) as functions of epochs. Overall accuracy (bottom) for train and validation sets over epochs.

600), and (3) non-smokers without COPD (N=200). At the initial study visit, participants (N=2,XXX) underwent non-contrast, full-lung CT scans acquired at total lung capacity (TLC) following a standardized imaging protocol.

As part of the SPIROMICS Repeatability sub-study, a subset of 100 Visit 1 participants underwent a second CT scan, with identical protocol, within 30 days of their initial visit. With repeated imaging on the same scanner, with the same protocol, and in a short interval, this dataset removes potential confounding due to disease progression or instrumentation changes, and allows for a study of the robustness of image-derived metrics of lung disease to variation in the CT acquisition alone - due to factors such as variable inspiratory effort.

MESA is a prospective cohort study that recruited 6,814 men and women in 2000-02 who were age 45-84 years and free of symptomatic cardiovascular disease at six clinical centers in Baltimore MD; Chicago, IL; Winston-Salem, NC; Los Angeles, CA; New York, NY; and St. Paul, MN. As part of Exam 5 of the MESA Lung Study (2010-2012), all participants (N=3,XXX) underwent full-lung CT following the SPIROMICS protocol.

Model training, validation and testing was performed on lung ROIs sampled from SPIROMICS Visit 1 participants as described below. ROIs sampled from MESA Lung Exam 5 are extracted and labeled by the final classifier as a measure of cross-cohort generalization of our model.

### A. ROI selection and pre-processing

For each subject in both SPIROMICS and MESA Lung exam 5, lung masks were generated using the APOLLO® software platform (VIDA Diagnostics, Inc., Coralville, Iowa). Ground-truth sLTP labels for all CT scans were obtained by segmentation following previously-published methods [10]. The Hounsfield intensity distribution over the lungs for each participant was normalized to [0,1]. ROIs of size 36x36x36 voxels are randomly sampled from all over the lung regions. An example of such ROI size is depicted in Figure 1. The ground-truth sLTP label of each ROI is by definition equal to the sLTP label at the ROI centroid.

### B. ROI sampling procedure and data augmentation

Input ROIs were randomly selected from 2,922 SPIROMICS Visit 1 participants under the following sampling criteria: (1) for each 3D sampled ROI with its centroid belonging to sLTP class $i$, 30% or more of the ROI volume must also be assigned to label $i$; (2) no two selected ROIs are allowed to have an overlap of more than 20% of their volume. Given the highly non-uniform distribution of selected ROIs in SPIROMICS Visit 1 (Figure 3a), we balanced the distribution of sampled ROIs by taking a random subsample of ROIs from each sLTP label to match the sample size of the least frequent class (sLTP 1). This led to a total of 8,640 sampled 3D ROIs for each of the 10 sLTP classes. A split ratio of 60:20:20 was used to form the train, validation, and test sets for the purpose of CNN training, validation, and testing (Figure 3b). We performed fourfold data augmentation on training samples by reflecting all ROIs along the $x$, $y$ and $z$ axes, for a final training

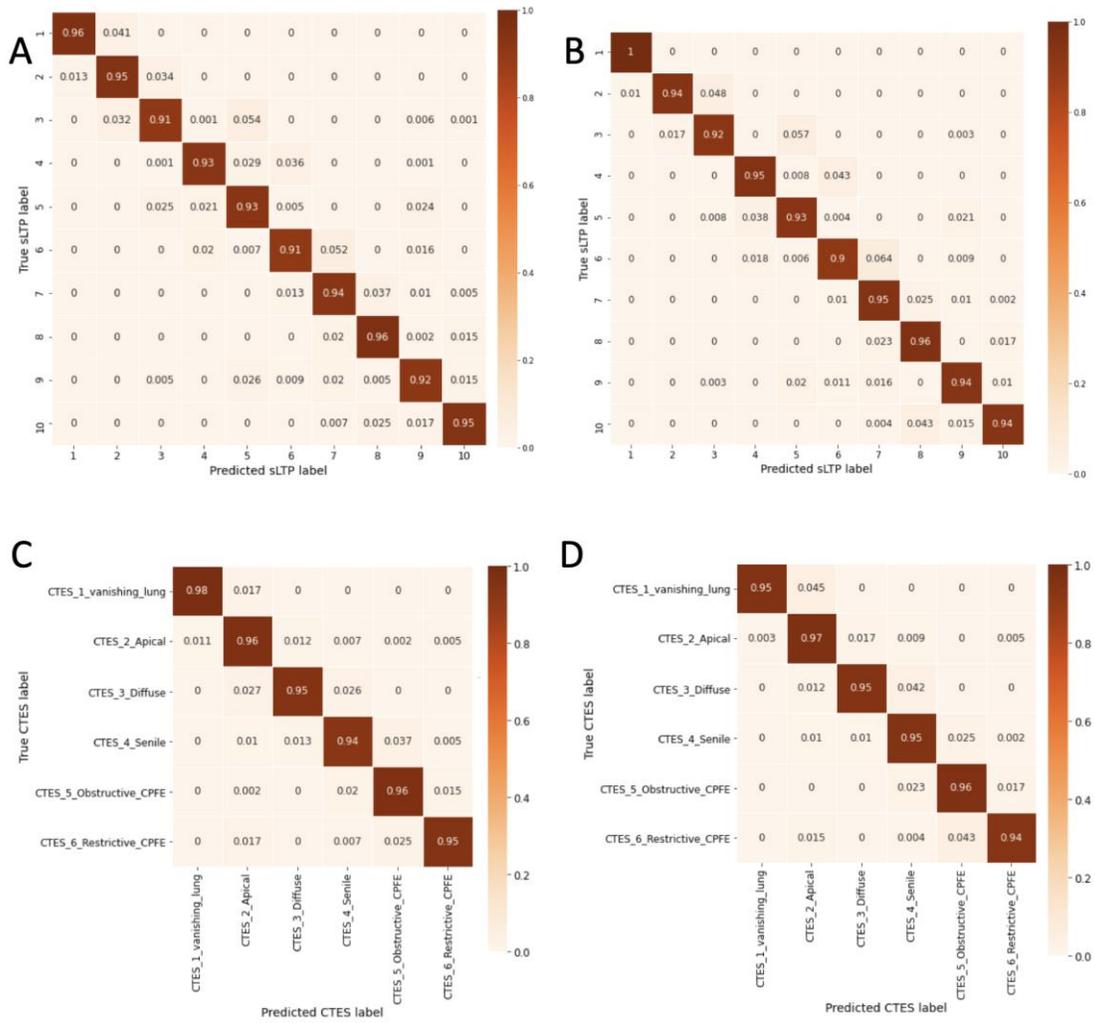

Figure 6. Confusion matrices for sLTP and CTES classification in the SPIROMICS Visit 1 and MESA Lung Exam 5 datasets: (A) SPIROMICS Visit 1 test set, sLTP classification; (B) MESA Lung Exam 5, sLTP classification; (C) SPIROMICS Visit 1 test set, CTES classification; (D) MESA Lung Exam 5, CTES classification

set of size 207,360 (20,736 per sLTP class) 3D ROIs.

Identical to the criteria for ROI sampling process in SPIROMICS, we sampled a total of 26,636 3D ROIs from 2,524 participants in MESA Lung exam 5. However, in order to evaluate the performance of the model with respect to the true distribution of sLTPs in this cohort, this time we did not carry out any balancing procedure and tested the model against the actual, nonuniform sampled sLTP distribution for MESA Lung Exam 5 (Figure 4).

### C. Scan-rescan Reproducibility Assessment

For all Repeatability participants with available CT (N = XX), ROIs are densely sampled from the Visit 1 and Repeatability scans by systematic uniform random sampling (SURS) over the respective lung masks; all emphysematous ROIs, i.e. those with a percentage of voxels below -950 HU intensity that exceeds the subject-specific upper limit of normal [27] are then assigned sLTP (CTES) labels using the existing pipeline [10] and our new deep model. The final sLTP (CTES) histogram is defined as the percentage of lung volume that is occupied by each sLTP (CTES).

## IV. EXPERIMENTAL RESULTS

### A. sLTP classification

The final trained model achieves top-1 sLTP classification accuracy of 98.15%, 94.16%, and 94.57% on the training, validation and test sets, respectively; the average inference time of our model was 1.49 ms per ROI. The test-set confusion matrix on SPIROMICS V1 (Figure 6A) shows top-1 classification accuracy between 91% and 96% for all ten sLTPs. Misclassification is greatest between adjacent sLTP indices; of the errors where the sLTP index differs by more than one, the most frequent classification errors include sLTPs 3 and 5, 5 and 9, and 4 and 6.

In order to assess the robustness of our model against different cohorts, we tested the model's performance against the MESA Lung Exam 5 dataset; there, the overall accuracy of our model was 93.85% (-0.72% drop relative to test-set performance in SPIROMICS), with top-1 accuracy for all sLTPs between 90% and 100% (Figure 6B).

Table 1. sLTP and CTES classification accuracy stratified by scanner model, on both the SPIROMICS Visit 1 test set and MESA Lung Exam 5 dataset. $N_{ROIs}$ denotes the number of distinct ROIs present in the sample for the given scanner model and cohort; $N_{participants}$ denotes the number of MESA/SPIROMICS participants imaged on the given scanner model.

| Scanner Model (SPIROMICS) | sLTP classification accuracy | CTES classification accuracy | $N_{ROIs}$ | $N_{participants}$ |
|---|---|---|---|---|
| Discovery CT750 HD | 93.61% | 95.59% | 4,380 | 486 |
| Discovery STE | 93.83% | 95.69% | 1,507 | 154 |
| LightSpeed VCT | 93.77% | 96.00% | 5,650 | 655 |
| Definition AS | 93.36% | 96.65% | 1,942 | 232 |
| Definition AS+ | 93.67% | 97.06% | 954 | 294 |
| Definition Flash | 94.35% | 96.90% | 902 | 157 |
| Sensation 64 | 92.22% | 95.28% | 784 | 55 |
| **Scanner Model (MESA Lung)** | **sLTP classification accuracy** | **CTES classification accuracy** | $N_{ROIs}$ | $N_{participants}$ |
| Discovery STE | 93.41% | 94.95% | 8,482 | 372 |
| LightSpeed VCT | 93.89% | 94.64% | 11,628 | 580 |
| Definition AS | 92.94% | 95.35% | 1,614 | 100 |
| Sensation 64 | 92.62% | 95.05% | 35774 | 1,469 |

Table 2. Participant-level reproducibility of sLTP histograms in SPIROMICS Repeatability, assessed by Pearson $R^2$ and Type 3 intraclass correlation coefficient (i.e., ICC(3,1)), for both the ground-truth and SE-CNN segmentation methods.

| sLTP index | Ground-truth $R^2$ | SE-CNN $R^2$ | Ground-truth ICC (3,1) | SE-CNN ICC(3,1) |
|---|---|---|---|---|
| 1 | 0.9979 [0.9945, 0.9992] | 0.9961 [0.9903, 0.9985] | 0.9927 [0.98, 1.0] | 0.9784 [0.95, 0.99] |
| 2 | 0.9576 [0.9151, 0.9790] | 0.9315 [0.8676, 0.9651] | 0.9747 [0.95, 0.99] | 0.9553 [0.91, 0.98] |
| 3 | 0.9779 [0.9607, 0.9876] | 0.9701 [0.9479, 0.9829] | 0.9864 [0.98, 0.99] | 0.9827 [0.97, 0.99] |
| 4 | 0.8503 [0.7656, 0.9061] | 0.8510 [0.7675, 0.9062] | 0.9205 [0.87, 0.95] | 0.9213 [0.87, 0.95] |
| 5 | 0.9853 [0.9758, 0.9911] | 0.9847 [0.9756, 0.9829] | 0.9924 [0.99, 1.0] | 0.9922 [0.99, 1.0] |
| 6 | 0.8792 [0.8177, 0.9210] | 0.8764 [0.8154, 0.9182] | 0.9377 [0.9, 0.96] | 0.9361 [0.9, 0.96] |
| 7 | 0.7032 [0.5773, 0.7975] | 0.6811 [0.5500, 0.7809] | 0.8385 [0.76, 0.89] | 0.8249 [0.74, 0.88] |
| 8 | 0.6543 [0.5154, 0.7616] | 0.6426 [0.5001, 0.7529] | 0.7615 [0.65, 0.84] | 0.7436 [0.63, 0.83] |
| 9 | 0.9519 [0.9258, 0.9690] | 0.9443 [0.9158, 0.9633] | 0.9755 [0.96, 0.98] | 0.9690 [0.95, 0.98] |
| 10 | 0.9381 [0.9067, 0.9592] | 0.9336 [0.8985, 0.9569] | 0.9863 [0.95, 0.98] | 0.9654 [0.95, 0.98] |

Table 3. Participant-level reproducibility of CTES histograms in SPIROMICS Repeatability, assessed by Pearson $R^2$ and Type 3 intraclass correlation coefficient, for both the ground-truth and SE-CNN segmentation methods.

| sLTP index | Ground-truth $R^2$ | SE-CNN $R^2$ | Ground-truth ICC (3,1) | SE-CNN ICC(3,1) |
|---|---|---|---|---|
| Vanishing | 0.9845 [0.9686, 0.9924] | 0.9828 [0.9660, 0.9914] | 0.9921 [0.98, 1.0] | 0.9914 [0.98, 1.0] |
| Apical | 0.9876 [0.9807, 0.9920] | 0.9854 [0.9777, 0.9904] | 0.9935 [0.99, 1.0] | 0.9918 [0.99, 0.99] |
| Diffuse | 0.8841 [0.8248, 0.9243] | 0.8822 [0.8238, 0.9221] | 0.9400 [0.91, 0.96] | 0.9391 [0.91, 0.96] |
| Senile | 0.7032 [0.5773, 0.7975] | 0.6811 [0.5500, 0.7809] | 0.8385 [0.76, 0.89] | 0.8249 [0.74, 0.88] |
| oCPFE | 0.6543 [0.5154, 0.7616] | 0.6426 [0.5001, 0.7529] | 0.7615 [0.65, 0.84] | 0.7436 [0.63, 0.83] |
| rCPFE | 0.9381 [0.9067, 0.9592] | 0.9336 [0.8985, 0.9569] | 0.9863 [0.95, 0.98] | 0.9654 [0.95, 0.98] |

### B. Performance on CTES classification

Visual inspection and statistical analysis of sLTPs led to the identification of six clinically-significant CTES: the vanishing-lung CTES is composed of 1 and 2, sLTPs 3,5, and 9 form the apical-bronchitic CTES, sLTPs 4 and 6 form the diffuse CTES, sLTP 7 forms the senile CTES, sLTP 8 forms the obstructive Combined Pulmonary Fibrosis/Emphysema (CPFE) CTES, and sLTP 10 forms restrictive CPFE [10]. In addition to classification performance on sLTPs, we also investigate model performance with respect to these broader clinical subphenotypes. CTES-level classification performance on the SPIROMICS test set (Figure 6C) shows 96.02% classification accuracy, with 95.10% overall accuracy observed in MESA Lung Exam 5 (Figure 6D); furthermore, for all CTES composed of *multiple* sLTPs in the SPIROMICS test set (and for the apical-bronchitic and diffuse subtypes in MESA Lung), the CTES-level classification accuracy is higher than that of any constituent sLTPs. In the limited cases where the model *does* make a classification error, therefore, it tends to preferentially assign an sLTP label that is both morphologically and functionally similar to the target. Classification accuracies for either sLTPs or CTES do not meaningfully differ between SPIROMICS Visit 1 and MESA Lung Exam 5.

### C. Performance for individual scanner types

In order to provide evidence that the model is capable of learning scanner-invariant features for sLTP classification, we examined overall sLTP and CTES classification accuracy, *stratified by scanner model*, for both the SPIROMICS test set and MESA Lung Exam 5 (Table 1). Due to the low frequency of some sLTPs (particularly sLTPs 1 and 2, known as the vanishing-lung CTES) in MESA Lung Exam 5 (Figure 4), for the scanner-level analysis in Table 2 we sampled an additional 30,900 ROIs from the MESA Lung Exam 5 scans, resulting in a total of 57,536 ROIs. sLTP classification accuracy falls between 92 and 95% for all scanner models in both SPIROMICS Visit 1 and MESA Lung Exam 5; similarly, classification accuracy across CTES is between approximately 95% and 97% for all scanner types in both cohorts.

### D. Scan-rescan Reproducibility

To be considered *robust* descriptors of emphysematous lung texture, sLTP and CTES labels must be highly reproducible on repeated CT imaging. For both the ground-truth and deep-learned sLTP classifiers, we compute the sLTP and CTES histograms for the SPIROMICS Visit 1 and SPIROMICS Repeatability CT scans of all Repeatability participants; we then compare the reproducibility of the two approaches by computing the Pearson $R^2$ and Type-3 intraclass correlation coefficient (ICC) between the baseline and follow-up histograms generated by each method. Our deep model and the prior pipeline show roughly equal robustness to repeat imaging as measured by either statistic: in the case of sLTP classification (Table 2), ICC(3,1) for the existing pipeline falls in the range [0.7615, 0.9927] across all sLTPs, while the deep model returns ICC(3,1) within [0.7436, 0.9922] over the same classes; the difference in intraclass correlation between the two segmentation methods never exceeds 0.0209 for any individual sLTP. We observe similar trends for the comparative Pearson $R^2$ across sLTPs, using both methods, and for ICC(3,1) and $R^2$ in the case of CTES classification (Table 3).

## V. Discussion

The overall sLTP and CTES accuracies and confusion matrices for SPIROMICS and MESA Lung cohorts (Figure 6) show minimal misclassification across subtypes, with test-set sLTP accuracy of 94.57% and 93.85% in SPIROMICS Visit 1 and MESA Lung Exam 5, respectively. Where misclassification does occur, we note that a significant proportion of errors at the sLTP level are attributable to mislabeling of ROIs of a given sLTP *as another member of the same CTES class*, i.e. the same emphysema subphenotype. This has the effect of improving overall performance of our model with respect to the final phenotyping with direct clinical relevance; moreover, the fact that these classification errors aggregate within CTES strongly suggests that even without any explicit penalties in the network loss, the texture representation learned by our model re-capitulates the data-reduction approach used to define CTES, further increasing confidence in the validity of our model.

Considering that we trained and tested our model on one cohort (*i.e.,* SPIROMICS) and evaluated it independently against another unseen cohort (MESA Lung), observing minimal change in classifier performance, our results demonstrate the ability of our model in learning cohort-invariant features for emphysema subtyping; furthermore, based on the per-scanner performance results summarized in Table 1 and Table 2, our model learns scanner-invariant features from the data that are informative for sLTP and CTES classification. Furthermore, while reproducibility of quantitative lung CT metrics at even a single visit is a significant challenge due to variations in breath-holding, patient positioning, and other practical concerns related to lung CT acquisition, our model shows non-inferior reproducibility on repeat imaging when compared to the ground-truth model. Considering cohort-invariant and scanner-invariant aspect of our model as well as its relatively fast ROI classification performance at the inference stage (average of 1.49 ms/ROI) makes the framework a quick and robust end-to-end classifier for emphysema learning and subtyping.

## VI. Conclusion

In this paper, a 3-D SE-CNN model was trained to classify 10 spatially-informed lung texture patterns (sLTPs) of emphysema, aggregated into six CT emphysema subtypes (CTES), on cubic ROIs extracted from emphysematous lung CT scans. By training, validating, and testing the model on the SPIROMICS dataset, we demonstrated the model's ability to classify the sLTPs and CTES with 94.57%, and 96.02% test set accuracies, respectively. On replication in CT scans from the MESA Lung study, our results showed the robustness of our model with respect to different cohorts with 93.85% and 95.10% sLTP and CTES classification accuracies, respectively. Finally, we reported the per scanner sLTP and CTES classification accuracies providing evidence that the model learns scanner-invariant features useful for accurate emphysema subtyping, and demonstrated scan-rescan reproducibility of our model equivalent to that of the ground-truth labels. Our findings overall indicate (1) that our SE-CNN model is able to reliably identify lung texture patterns (sLTPs) and emphysema subphenotypes (CTES) with well over 90% and 95% accuracy, respectively; (2) that this performance is independnt of the cohort used to evaluate our model, independent of the CT scanner used, and robust to natural variability introduced by the image acquisition process.

## Acknowledgments

The authors thank the SPIROMICS participants and participating physicians, investigators and staff for making this research possible. More information about the study and how to access SPIROMICS data is available at www.spiromics.org. The authors would like to acknowledge the University of North Carolina at Chapel Hill BioSpecimen Processing Facility for sample processing, storage, and sample disbursements

(http://bsp.web.unc.edu/). We would like to acknowledge the following current and former investigators of the SPIROMICS sites and reading centers: Neil E Alexis, MD; Wayne H Anderson, PhD; Mehrdad Arjomandi, MD; Igor Barjaktarevic, MD, PhD; R Graham Barr, MD, DrPH; Patricia Basta, PhD; Lori A Bateman, MSc; Surya P Bhatt, MD; Eugene R Bleecker, MD; Richard C Boucher, MD; Russell P Bowler, MD, PhD; Stephanie A Christenson, MD; Alejandro P Comellas, MD; Christopher B Cooper, MD, PhD; David J Couper, PhD; Gerard J Criner, MD; Ronald G Crystal, MD; Jeffrey L Curtis, MD; Claire M Doerschuk, MD; Mark T Dransfield, MD; Brad Drummond, MD; Christine M Freeman, PhD; Craig Galban, PhD; MeiLan K Han, MD, MS; Nadia N Hansel, MD, MPH; Annette T Hastie, PhD; Eric A Hoffman, PhD; Yvonne Huang, MD; Robert J Kaner, MD; Richard E Kanner, MD; Eric C Kleerup, MD; Jerry A Krishnan, MD, PhD; Lisa M LaVange, PhD; Stephen C Lazarus, MD; Fernando J Martinez, MD, MS; Deborah A Meyers, PhD; Wendy C Moore, MD; John D Newell Jr, MD; Robert Paine, III, MD; Laura Paulin, MD, MHS; Stephen P Peters, MD, PhD; Cheryl Pirozzi, MD; Nirupama Putcha, MD, MHS; Elizabeth C Oelsner, MD, MPH; Wanda K O'Neal, PhD; Victor E Ortega, MD, PhD; Sanjeev Raman, MBBS, MD; Stephen I. Rennard, MD; Donald P Tashkin, MD; J Michael Wells, MD; Robert A Wise, MD; and Prescott G Woodruff, MD, MPH. The project officers from the Lung Division of the National Heart, Lung, and Blood Institute were Lisa Postow, PhD, and Lisa Viviano, BSN; SPIROMICS was supported by contracts from the NIH/NHLBI (HHSN268200900013C,HHSN268200900014C,HHSN268200900015C, HHSN268200900016C, HHSN268200900017C, HHSN268200900018C,HHSN268200900019C,HHSN268200900020C), grants from the NIH/NHLBI (U01 HL137880 and U24 HL141762), and supplemented by contributions made through the Foundation for the NIH and the COPD Foundation from AstraZeneca/MedImmune; Bayer; Bellerophon Therapeutics; Boehringer-Ingelheim Pharmaceuticals, Inc.; Chiesi Farmaceutici S.p.A.; Forest Research Institute, Inc.; GlaxoSmithKline; Grifols Therapeutics, Inc.; Ikaria, Inc.; Novartis Pharmaceuticals Corporation; Nycomed GmbH; ProterixBio; Regeneron Pharmaceuticals, Inc.; Sanofi; Sunovion; Takeda Pharmaceutical Company; and Theravance Biopharma and Mylan.